\documentclass[journal]{IEEEtran}
\usepackage{graphics} 
\usepackage{epsfig}
\usepackage[utf8]{inputenc}
\usepackage{nomencl}
\usepackage[mathscr]{euscript}
\usepackage{multirow}
\makenomenclature

\usepackage{times} 
\usepackage{graphicx}
\usepackage{float}
\usepackage{algorithm}
\usepackage{algorithmic}

\usepackage{tikz}
\usetikzlibrary{shapes,arrows}

\usepackage{verbatim}
\usepackage{amsmath} 
\usepackage{amssymb}  
\usepackage{amsmath}
\usepackage{amssymb}
\usepackage{mathtools}
\usepackage{mathrsfs}

\usepackage{graphicx}              
\usepackage{multirow}
\usepackage{amsthm}
\usepackage{booktabs}
\usepackage{colortbl}
\graphicspath{{DTL_Figs/}}

\DeclareSymbolFont{rsfs}{U}{rsfs}{m}{n}
\DeclareSymbolFontAlphabet{\mathscrsfs}{rsfs}

\title{Wasserstein Distance based Deep Adversarial Transfer Learning for Intelligent Fault Diagnosis}
\author{Cheng Cheng, Beitong Zhou, Guijun Ma, Dongrui Wu and Ye Yuan
\thanks{Cheng Cheng, Beitong Zhou, Dongrui Wu and Ye Yuan are with School of Automation, Huazhong University of Science and Technology, Wuhan, China, 430074. Ye Yuan is also with State Key Lab of Digital Manufacturing Equipment and Technology, Wuhan, China, 430074. Guijun Ma is with School of Mechanical Science and Engineering, Huazhong University of Science and Technology, Wuhan, China, 430074. For correspondence, contact Prof. Ye Yuan (yye@hust.edu.cn). }
}

\begin{document}

\maketitle

\begin{abstract}
The demand of artificial intelligent adoption for condition based maintenance strategy is astonishingly increased over the past few years. Intelligent fault diagnosis is one critical topic of maintenance solution for mechanical systems. Deep learning models, such as convolutional neural networks (CNNs), have been successfully applied to fault diagnosis tasks for machinery systems, and achieved promising results. However, for diverse working conditions in industry, deep learning suffers two difficulties: one is that the well-defined (source domain) and new (target domain) datasets are with different feature distributions; and another one is the fact that insufficient or no labelled data in target domain significantly reduce the accuracy of fault diagnosis. As a novel idea, deep transfer learning (DTL) is created to perform learning in the target domain by leveraging information from relevant source domain. Inspired by Wasserstein distance of optimal transport, in this paper, we propose a novel DTL approach to intelligent fault diagnosis, namely Wasserstein Distance based Deep Transfer Learning (WD-DTL), to learn domain feature representations (generated by a CNN based feature extractor) and to minimize the distributions between the source and target domains through adversarial training. The effectiveness of the proposed WD-DTL is verified through 3 transfer scenarios and 16 transfer fault diagnosis experiments of both unsupervised and supervised (with insufficient labeled data) learning. We also provide comprehensive analysis on the network visualization of those transfer tasks.
\end{abstract}


\section{Introduction}
\label{sec:Intro}
\IEEEPARstart{F}{ault} diagnosis aims to isolate faults on defective systems by monitoring and analyzing machine status using acquired measurements and other information, which requires experienced experts with a high skill set. This drives the demand of artificial intelligent techniques to make fault diagnosis decisions. The deployment of a real-time fault diagnosis framework allows the maintenance team to act in advance to replace or fix the affected components, thus, improving production efficiency and guarantee operational safety.


Over the past decade, many advanced signal processing and machine learning techniques have been used for fault diagnosis. Signal processing techniques such as wavelet \cite{bae2017condition} and Hilbert-Huang transform \cite{samanta2003artificial} are adopted for feature extraction from faulty vibration signals, and machine learning models are then applied to automate the fault diagnosis procedure. In last few years, deep learning models, such as deep belief networks (DBN) \cite{tamilselvan2013failure}, sparse auto-encoder \cite{chen2017multisensor}, and especially convolutional neural networks (CNN) \cite{hu2017intelligent}, have shown superior fitting and learning ability in fault diagnosis tasks over ruled-based and model-based methods. However, the above stated deep learning approaches suffer two difficulties: 1) Most of the approaches work well under a same hypothesis: the datasets for source domain and target domain tasks are required to be identically distributed. Thus, the adaptability of the pre-trained network is limited when facing new diagnosis task, where the different operational conditions and physical characteristics of the new task might cause distribution difference between the new dataset (target dataset) and the original dataset (source dataset). As a result, for a new fault diagnosis task, the deep learning model is commonly reconstructed from scratch, which results in the waste of computational resources and training time; 2) Insufficient labeled or unlabeled data in target domain is another common problem. In real industry situations, for a new diagnosis task, it is extremely difficult to collect sufficient typical samples to re-build a large-scale and high-quality dataset to train a network. 


Deep transfer learning (DTL) \cite{guo2018deep, sun2018deep} aims to perform learning in a target domain (with insufficient labeled or unlabeled data) by leveraging knowledge from relevant source domains (with sufficient labeled data), saving much expenditure on reconstructing a new fault diagnosis model from scratch and recollecting sufficient diagnosis labeled samples. Many successful approaches to DTL has been seen in various fields, including pattern recognition \cite{gopalan2011domain}, image classification \cite{patel2015visual}, and speech recognition \cite{deng2014autoencoder}.

Solutions to DTL can be roughly classified into three categories: instances-based DTL, network-based DTL, and mapping-based DTL.  Instances-based DTL reweighs/subsamples a group of instances from the source domain to match the distributions in the target domain. Network-based DTL crops out partial of the network pre-trained in the source domain, which is transferred to be a part of target network for a relevant new task; see \cite{yao2010boosting, yosinski2014transferable} for recent examples of instances-based and network-based DTL, respectively. 
However, above approaches are not capable of learning a latent representation from the deep architecture. Mapping-based DTL, compared with other approaches to adapting deep models, has shown excellent properties through finding a common latent space, where the feature representations for source and target domains are invariant.  Tzeng \textit{et. al} \cite{tzeng2014deep} proposed a CNN architecture based network for domain adaptation, which introducing an adaptation layer to learn the feature representations. Maximum mean discrepancy (MMD) metric is used as an additional loss for the overall structure to compute the distribution distance with respect to a particular representation, which helps to select the depth and width of the architecture as well as to regulate the loss function during fine-tuning. Later, in \cite{long2015learning} and \cite{long2017deep}, MMD was extended to multiple kernel variance MMD (MK-MMD) and joint MMD (JMMD) for better domain adaptation performance. However, the limitation of MMD method for domain adaptation is that the computational cost of MMD is quadratically increased with large mount of samples when calculating the Integral Probability Metrics (IPMs) \cite{gretton2012kernel}. Recently, Ajovsky \textit{et al.} \cite{arjovsky2017wasserstein} indicate that Wasserstein distance can be a new direction to find better distribution mapping. Compared with other popular probability distances and divergences, such as Kullback-Leibler (KL) divergence and Jensen-Shannon (JS) divergence, \cite{arjovsky2017wasserstein} demonstrated that Wasserstein distance is a more sensible cost function when learning distributions supported by low dimensional manifolds. Later on, \cite{gulrajani2017improved} and \cite{shen2018wasserstein} proposed a new gradient penalty term for domain critic parameters to solve the gradient vanishing or exploding problems in \cite{arjovsky2017wasserstein}. Hence, the essence of our proposed approach is to adopt the Wasserstein distance to train a DTL model for intelligent fault diagnosis problem which seeks to minimize the distributions between source domain and target domains. Our motivation of this work is to figure out how Wasserstein distance behaves in transfer learning due to its excellent performance in generative adversarial network (GAN). 


This paper concerns the problem of DTL modeling to explore the transferable features of fault diagnosis under different operating conditions, including different motor speeds, and different sensor locations. Firstly, in source domain, a base CNN model is trained with sufficient data. Then, we build a Wasserstein distance based DTL (WD-DTL) to learn invariant features between source and target domains. A neural network is introduced (denoted by domain critic) to calculate the empirical Wasserstein distance by maximizing domain critic loss. After this procedure, a discriminator is introduced to optimize the CNN-based feature extractor parameters by minimizing the estimated empirical Wasserstein distance.  Through the above adversarial learning process, the transferable features from a source domain where faulty labels are known can be brought to diagnose a new but relevant diagnosis task without any labeled sample. To our best knowledge, this is the first work adopts the Wasserstein distance to CNN for measuring the domain distance in fault diagnosis problems. Experimental results, through 16 transfer tasks, demonstrate the effectiveness of the distance measurement method and the proposed DTL model. This paper makes the following contributions:

\begin{enumerate}
\item Wasserstein distance is used as the distance measurement of domains in fault diagnosis problems to explore better distribution mapping. Mapping features are extracted by a pre-trained CNN based feature extractor.
\item The proposed WD-DTL framework could perform both  unsupervised and supervised transfer tasks. Consequently, for a new diagnosis task, this is a novel approach which could contribute to solve both unlabeled and insufficient labeled data in real industry applications. Extensive experiments will be conducted to support this statement.
\item The versatility of our WD-DTL approach is demonstrated with transfer learning experiments, in terms of 3 different transfer scenarios and 16 transfer tasks in total. To emphasize, the proposed WD-DTL approach surpass the existing transfer learning network DAN with MK-MMD in almost all transfer tasks.
\end{enumerate}


This paper is organized as follows. Section \ref{sec:related_work} reviews related works including CNN for fault diagnosis and transfer learning. Section \ref{sec:method} proposes our intelligent fault diagnosis framework by using transfer learning method. Experiment results and comparison are given in Section \ref{sec:tests}. Finally, conclusion and future work are drawn in Section \ref{sec:conclusion}.

The following notations will be used throughout this work: the symbol $\mathbb{R}$ is the real number set, and the symbol $\mathbb{Z}$ is the positive integer set. $(\cdot)^s$ and $(\cdot)^t$ represent the source and target domain information respectively.

\section{Related works}
\label{sec:related_work}
In this section, some related work on intelligent fault diagnosis as well as CNN architecture are provided, and followed by a brief introduction associated with transfer learning and Wasserstein distance.

\subsection{Convolutional Neural Networks}
\label{sec:Fault_diagnosis}
As the most well-known model in deep learning, in recent years, CNN dominates the recognition and detection problems in computer vision domain. The initial CNN architecture was proposed by LeCun \textit{et al.} in works \cite{lecun1990handwritten} and \cite{lecun1989backpropagation}, which was inspired by Wiesel and Hubel's research works in cat recognition \cite{hubel1959receptive}. Main characteristics of CNN are local connections, shared weights, and local pooling \cite{saxe2011random}. The first two characteristics indicate the CNN model require less parameters to detect local information of visual patterns than multilayer perceptron, while the last characteristic offers shift invariance to the network. Typically, 1-D CNN will be employed to this work to solve the bearing fault diagnosis problem, which has been widely used with great success in the study of speech recognition and document reading tasks. 

In this work, a 1-D CNN model, as a base model, will be pre-trained in source domain. The CNN extract and learn characteristics of the task by stacking a series of layers with repeated components, including convolutional layers (with activation function), pooling layers, and fully connected layers (with an output classification layer) \cite{lecun2015deep}. A typical CNN architecture is fed to a 1-D input layer to accept source domain signal, convolutional layers with rectified linear unit (ReLU) activation functions are followed for feature extraction, max pooling layers are used to down-sampling data size, and a fully connected layer combined with a softmax function is finally connected for classification (with pre-defined labels). To minimize the loss function, model parameters are tuned using Backpropagation algorithm \cite{vogl1988accelerating} based on stochastic gradient descent (SGD) optimizer, until the predefined maximum number of iterations is reached. More details and expressions of each layer for the bearing fault diagnosis task will be explained in Section \ref{sec:method}.

\subsection{Transfer learning}
\label{sec:DTL}

Transfer learning can be a novel tool to solve the basic problem of unlabeled and insufficient data under diverse operating conditions in target domain of mechanical systems, by utilizing the knowledge from source domain to improve the target domain learning performance. Some notations and definitions of transfer learning used in this work are first presented.

To begin with, we define a domain and a task respectively. Given a domain $\mathscr{D}$ in transfer learning defined as $\mathscr{D}=\{\mathcal{X},\mathbb{P}(X)\}$, where $\mathbb{P}(X)$ represents a marginal probability distribution of a feature space $\mathcal{X}$. Given predefined source and target domain datasets $X^s$ and $X^t$, we have $X^s, X^t\in\mathcal{X}$. If $X^s\neq X^s$ and/or $\mathbb{P}^s(X^s)\neq\mathbb{P}^t(X^t)$, two domains $\mathscr{D}^s$ and $\mathscr{D}^t$ are with different distribution.

In the meantime, a task $\mathscr{T}$ in transfer learning is defined as $\mathscr{T}=\{\mathcal{Y},r(X)\}$, where $\mathcal{Y}$ represents a label space and $r(X)$ is a predictive function and $r(X)=\mathbb{P}(Y|X)$ is a conditional probability function. Since the classification categories are the same, source and target domains have the same label space, $\mathcal{Y}^s=\mathcal{Y}^t$. Then, we give the definition of transfer learning.

\textbf{Definition 1. (\textit{Transfer learning})} Transfer learning is proposed with the aim to learn a prediction function $r(X):X\longrightarrow Y$ for a learning task $\mathscr{T}^t$ by leveraging knowledge from source domain $\mathscr{D}^s$ and $\mathscr{T}^s$, where $\mathscr{D}^s\neq\mathscr{D}^t$ or $\mathscr{T}^s\neq\mathscr{T}^t$. In most of the cases, $\mathscr{D}^s$ contains a much larger dataset than $\mathscr{D}^t$ (i.e., cardinality of $\mathscr{D}^s$ is larger than that of $\mathscr{D}^t$). 

\subsection{Wasserstein distance}
\label{sec:W_distance}

Wasserstein distance is recently proposed by researchers \cite{arjovsky2017wasserstein} to tackle the training difficulty of generative adversarial networks (GAN) when facing discontinuous mapping problem of other distances and divergences in the generator, such as Total Variation (TV) distance and Kullback-Leibler (KL) divergence. As an promising way to measure the distance between two distributions for GAN training, Wasserstein distance could be applied to DTL for domain adaptation. 

Given a compact metric set $\mathcal{H}$, $Prob(\mathcal{H})$ represents the space of probability measures on set $\mathcal{H}$. Wasserstein-1 distance (also called $\textit{Earth-Mover}$ distance) is defined between two distributions $\mathbb{P}^s$, $\mathbb{P}^t$ $\in Prob(\mathcal{H})$:

\begin{equation}
  W(\mathbb{P}^s,\mathbb{P}^t)=\inf \limits_{\mu\in \Pi(\mathbb{P}^s,\mathbb{P}^t)}\mathbb{E}_{(h^s,h^t)\sim \mu}[\parallel h^s-h^t \parallel]
  \label{eq:ws_1}
\end{equation}
where $\mu$ is a joint probability distribution and $\Pi(\mathbb{P}^s,\mathbb{P}^t)$ denotes the set $\mathcal{H}\times \mathcal{H}$ of all joint distributions $\mu(h^s,h^t)$ whose marginals are $\mathbb{P}^s$ and $\mathbb{P}^t$ respectively. Wasserstein-1 distance can be viewed as a optimal transport problem, it is aims to find an optimal transport plan $\mu(h^s,h^t)$. Intuitively, $\mu(h^s,h^t)$ indicates how much of `mass' randomly transported from one place $h^s$ over the domain of $h^t$, with the aim of transporting the distribution $\mathbb{P}^s$ into the distribution $\mathbb{P}^t$. Hence, Wasserstein-1 distance is the optimal transport plan with the lowest transport cost.

\section{Wasserstein Distance based Deep Transfer Learning (WD-DTL)}
\label{sec:method}

\subsection{Problem formulation}
\label{sec:Prob}

Since it is difficult to retrofit enough sensors in packaged equipment and industry labeling often requires expensive human efforts for mechanical systems, the challenge of domain adaptation is that there is no or limited labeled high-quality data can be collected in target domain. For this reason, supervised domain adaptation approach by fine-tuning the pre-trained architecture to fit the new classification problem in target domain is not feasible. To solve this problem, many existing domain adaptation frameworks \cite{gretton2012kernel, long2015learning} using MMD to learn the invariant domain representations, which minimizing the target loss by the source loss with an additional maximum mean discrepancy metric. Our proposed approach WD-DTL is a promising alternative for domain adaptation by using the Wasserstein distance, which has been demonstrated with gradient superiority than MMD \cite{arjovsky2017wasserstein}, to minimize the distributions between source domain and target domain. Although Wasserstein distance with MLP has been seen in few domain adaptation works in image classification tasks, to date there is no attempt to adopt this technique into industry or manufacturing and there is no attempt to enhance this technique in deep neural networks. It also has to be noted that we propose to use the CNN architecture to generate features for measuring the Wasserstein distance in both domains, meanwhile, the excellent local feature detection ability of CNN in manufacturing has been explored in work \cite{cheng2018online}. The problem of this work is formulated as follow:

The DTL with domain adaptation for fault diagnosis is an unsupervised problem, thus, we first define a source domain dataset with labels $y^s_i$ by $X^s=\{(x^s_i,y^s_i)\}^{N^s}_{i=1}$ with $N^s\in \mathbb{R}$ number of samples in the source domain $\mathscr{D}^s$. In the meantime, an unlabeled target domain dataset $X^t=\{x^t_i\}^{N^t}_{i=1}$ is defined in the target domain $\mathscr{D}^t$. In most cases, source domain samples are sufficient enough to learn an accurate CNN classier and with much larger data size than the target domain, which means $N^s \gg N^t$. It is also noted that data in source and target domains share the same feature space ($X^s, X^t\in \mathcal{X}$) but with different marginal distributions ($\mathbb{P}^{s}(X^s)\neq \mathbb{P}^{t}(X^t)$).

The objective of this work is to construct a transferable framework, named WD-DTL, for the target task $\mathscr{T}^t$ to minimize target classification error $E^t\%=\textup{Pr}_{(x^t,y^t)\sim \mathscr{D}^t}[r(X^t)\neq y^t]$, with the help of the knowledge from source domain task $\mathscr{T}^s$ and the implementation of Wasserstein distance for domain adaptation.  

The algorithm of WD-DTL will be trained by three iterative steps: a CNN-based feature extractor will be described in Section \ref{sec:FE}; domain adaptation using Wasserstein distance will be explained in Section \ref{sec:DA}; and finally a discriminator for classification in Section \ref{sec:Cf_discriminator}.


\subsection{CNN based feature extractor}
\label{sec:FE}
First of all, we propose to use CNN to train the domain data. A CNN model is pre-trained with source domain labelled dataset $X^s$:

\textit{Convolution layer} involves a filter $w\in \mathbb{R}^k$ and a bias $b\in \mathbb{R}$, which are applied to a filter size of $k$ for calculating a new feature. An output feature $v_i$ is obtained through the filter $w$ and a non-linear \textit{activation function} $\Gamma$ with the following expression:

\begin{equation}
v_i=\Gamma \,(w\ast u_j+b),
\label{eq:sigment_conv}
\end{equation}
where $u_j\in \mathbb{R}^{1\times k}$ is the input data representing $j$-th sub-vector of the source domain dataset $X^s$. `$\ast$' denotes the convolution operation. The non-linear activation function, such as hyperbolic tangent (tanh) or rectified linear unit (ReLu), is applied to reduce the risk of vanishing gradient which may impact the convergence of the optimization. Hence, the \textit{feature map} is defined as ${\bf{v}}=[v_1,v_2,\ldots,v_L]$, where $L=(pN-s)/I_{cv}+1$ is the number of features and $I_{cv}\in \mathbb{Z}$ is the stride for convolution.

\textit{Max pooling layer} is then applied over the feature map to extract the maximum feature values $\widehat{v_i}=\max^{}\limits_{\gamma=1,\ldots,\beta}\,\,v_{\gamma+(i-1)I_{pl}}$ corresponding to its filter size $\beta$ and the stride size $I_{pl}$ for max pooling. The idea is to capture the maximum features over disjoint regions. Consequently, the features within the small window are similar and therefore illustrating the most important property of CNN.

By stacking multiple layers described above (with varying filter size), a multi-layer structure is constructed for feature description. The output features of the multi-layer structure is flattened and pass to fully-connected layers for classification, resulting in probability-distributed final outputs $\tilde{y}^s_i$  over labels. For the pre-trained CNN in source domain, \textit{Softmax} function \cite{girshick2015fast} is selected for classification over the final feature map. 

To compute the difference between the predicted label, $\tilde{y}^s_i$, and the ground truth, $y^s_i$, in source domain, cross-entropy function $l_c$ is used to compute the loss:
\begin{equation}
l_c=\frac{1}{N^s}\sum_{i=1}^{N}-y^s_i\, \textup{log} \tilde{y}^s_i-(1-y^s_i) \,\textup{log} (1-\tilde{y}^s_i).
\label{eq:lossfx}
\end{equation}

\subsection{Domain adaptation via Wasserstein distance}
\label{sec:DA}
Transferable features of the target domain with unlabeled data or insufficient labeled data can be 
directly obtained by the pre-trained accurate CNN feature extractor of the last subsection. The next problem is to solve the distribution difference between the source and target datasets. To tackle this problem, we utilize Wasserstein-1 distance to learn invariant feature representations in a common latent space between two different feature distributions through adversarial training.   
 
                                                                                                                                                                                                                                                                                                                                                                                                                                                                                                                                                                                                                                                                                                                                                                                                                                                                                                                                                                                                                                                                                                                                                                                                                                                                                                                                                                                                                                                                                                                                                                                                                                                                                              The network structure before fully-connected layer of pre-trained CNN model is used as the feature extractor to learn the invariant feature representations from both domains. Given two mini-batch of instances $\{x^s\}^{n}_{i=1}$ and $\{x^t\}^{n}_{i=1}$ from $X^s$ and $X^t$ for $n<N^s \,\, \textup{and}\,\, N^t$. Both instances are passed through a parameter function $r_f$: $\mathcal{X}\rightarrow \mathcal{H}$ (i.e., feature extractor) with corresponding network parameter $\theta_f$ that directly generate source features $h^s=r_f(x^s)$ and target features $h^t=r_f(x^t)$. Let $\mathbb{P}^s$ and $\mathbb{P}^t$ be the distribution of $h^s$ and $h^t$ respectively. 
 
 The aim of domain adaptation via Wasserstein distance \cite{arjovsky2017wasserstein} is to optimize the parameter $\theta_f$ to reduce the distance between distributions $\mathbb{P}^s$ and $\mathbb{P}^t$. We introduce a domain critic learns a solution $r_c$: $\mathcal{H}\rightarrow\mathbb{R}$ that maps the source and target features to a real number, with corresponding parameters $\theta_c$. However, the equation of the \textit{infimum} in Eq. \eqref{eq:ws_1} is highly intractable to handle directly. Thanks to the Kantorovich-Rubinstein duality \cite{villani2008optimal}, the Wasserstein-1 distance can be computed by

\begin{equation}
  W(\mathbb{P}^s,\mathbb{P}^t)=
 \sup \limits_{\parallel r_c \parallel\leq 1}\mathbb{E}_{h^s\sim\mathbb{P}^s}[r_c(h^s)]-\mathbb{E}_{h^s\sim\mathbb{P}^t}[r_c(h^s)]
  \label{eq:ws_2}
\end{equation}
where the \textit{supermum} is over all the 1-Lipschitz functions $r_c$: $\mathcal{H}\rightarrow\mathbb{R}$. The empirical Wasserstein-1 distance  can be approximately computed as follow:
 
 \begin{equation}
 l_{wd}=\frac{1}{N^s}\sum \limits_{x^s\in X^s} r_c(r_f(x^s))-\frac{1}{N^t}\sum \limits_{x^t\in X^t} r_c(r_f(x^t)).
  \label{eq:EWD}
\end{equation}
where $l_{wd}$ denotes the domain critic loss between the source data $X^s$ and the target data $X^t$

Now comes to the optimization problem that find the maximum of Eq. \eqref{eq:EWD} while enforcing the Lipschitz constraint. Arjovsky \textit{et al.} \cite{arjovsky2017wasserstein} proposed a weight clipping method after each gradient update to force the parameters $\theta_c$ inside a compact space. However, this method is time consuming when clipping parameter is large and might result in vanishing gradients when the number of layers is set too big. To solve this problem, \cite{shen2018wasserstein, gulrajani2017improved} suggest to incorporate a gradient penalty $l_{grad}=(\parallel \nabla_{\textbf{h}}r_c(\textbf{h})\parallel_2-1)^2$ to train the domain critic with respects to parameters $\theta_c$, where the feature representations $\textbf{h}$ consist of the generated source and target domain features (i.e., $h^s$ and $h^t$), as well as points $h^r$ which are randomly selected along the straight line between $h^s$ and $h^t$ pairs. 


As the fact that the Wasserstein-1 distance is differentiable and continuous almost everywhere, we here to train the critic till optimally by solving the following optimization problem:

\begin{equation}
  \max \limits_{\theta_c}\{l_{wd}-\rho l_{grad}\}
\end{equation}
where $\rho$ is the balancing coefficient.

\subsection{Classification with discriminator}
\label{sec:Cf_discriminator}
The above Section \ref{sec:DA} proposed an unsupervised feature learning for domain adaptation, which may cause the learned feature representations in both domains are not discriminative enough. As stated in Section \ref{sec:Prob}, our final objective is to develop an accurate classifier, WD-DTL, for target domain $\mathscr{D}^t$, which requires to incorporate the labelled supervised learning of source domain data (and target domain if avaliable) into the invariant feature learning problem. A discriminator \cite{ganin2016domain} (with two fully-connected layers) is then employed into the representation learning approaches to further reduce the distance between source and target feature distributions. In this step, parameters of domain critic $\theta_c$ are the ones trained in Section \ref{sec:DA}, while the parameters $\theta_f$ will be modified to optimize the minimum operator.

Now the final objective function can be expressed in terms of the cross-entropy loss $l_c$ of the discriminator according to Eq. \eqref{eq:lossfx} and the empirical Wasserstein distance $l_{wd}$ which associated with domain discrepancy, i.e:

\begin{equation}
\min \limits_{\theta_d,\theta_f}\lbrace l_c+\lambda \max \limits_{\theta_c}[l_{wd}-\rho l_{grad}] \rbrace
  \label{eq:total_obj}
\end{equation}
where $\theta_d$ denotes the parameters for the discriminator and $\lambda$ is the hyper-parameter that determines the extent of domain confusion. We omit the gradient penalty $l_{grad}$ (i.e., set $\rho$ equal to 0) when optimizing the minimum operator as it should not affect the representation learning process.

\subsection{WD-DTL Approach}
\label{sec:approach}
\begin{figure*}[hbt!]
\centering
  \includegraphics[width=1.8\columnwidth]{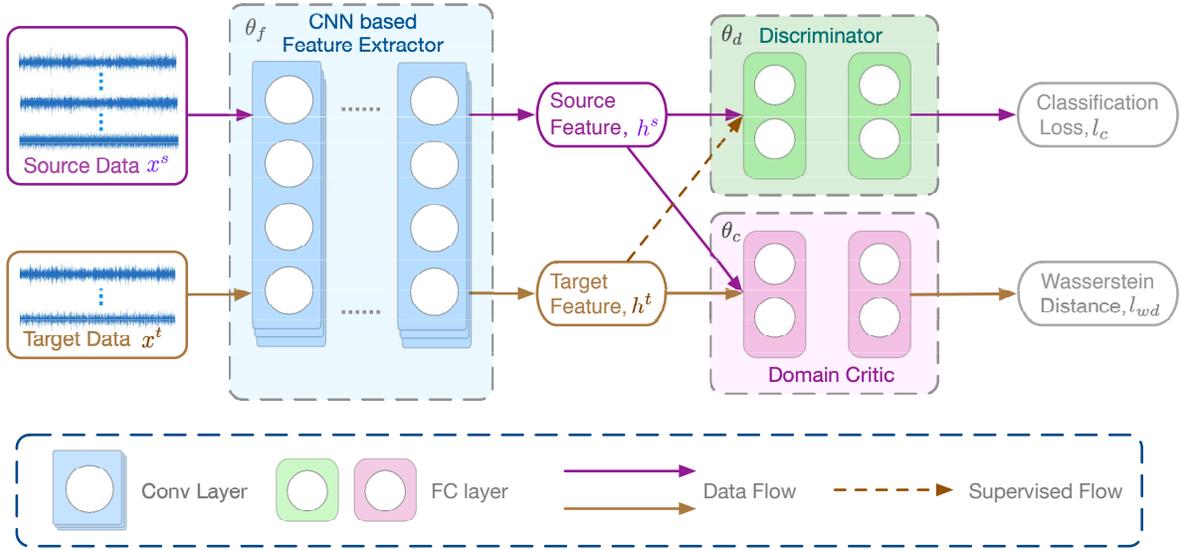}
  \caption{WD-DTL framework of the fault diagnosis, which is comprised of three sub networks: a CNN based feature extractor, a domain critic for learning feature representations via Wasserstein distance, and a discriminator for classification. Dashed yellow line shows that this framework is also can be used for supervised transfer learning.}
  \label{fig:WD-DTL}
\end{figure*}

Hence, the overall framework of intelligent fault diagnosis approach in this work is illustrated in Fig. \ref{fig:WD-DTL} and a detailed algorithm is summarized in Algorithm \ref{code:alg1}.

\begin{algorithm}[h]
\caption{Training procedure of WD-DTL.}
\begin{algorithmic}[1]
\label{code:alg1}
\REQUIRE source and target dataset: $X^s$ and $X^t$; the learning rate for domain critic: $\alpha_1$; the learning rate for discriminator and feature learning: $\alpha_2$; minibatch size for source and target datasets: $n$; critic training step: $C$; balance coefficients: $\rho$ and $\lambda$.
\ENSURE initial CNN based feature extractor parameters: $\theta_f$; initial domain critic parameters: $\theta_c$; initial discriminator parameters: $\theta_d$.
\WHILE {$\theta_f$, $\theta_c$, and $\theta_d$ has not converged }
\STATE Sample $\{x^s_i,y^s_i\}^{n}_{i=1}$, a batch from source dataset $X^s$.
\STATE Sample $\{x^t_i\}^{n}_{i=1}$, a batch from target dataset $X^t$.
\FOR{$i=0$ to $C$}
\STATE $h^s \leftarrow r_f(x^s)\sim \mathbb{P}^s$ , $h^t \leftarrow r_f(x^t)\sim \mathbb{P}^t$
\STATE Sample $h^r$ from $h^s$ and $h^t$ pairs, 
\STATE {\bf{h}} $\leftarrow \{h^s,h^t,h^r\}$
\STATE $l_{grad}\leftarrow  (\parallel \nabla_{\textbf{h}}r_c(\textbf{h})\parallel_2-1)^2$
\STATE $\theta_c \leftarrow \theta_c + \alpha_1 \nabla_{\theta_c}[l_{wd}(x^s,x^t)-\rho l_{grad}(\textbf{h})]$
\ENDFOR
\STATE $\theta_d \leftarrow \theta_d - \alpha_2 \nabla_{\theta_d} l_c(x^s,y^s)$
\STATE $\theta_f \leftarrow \theta_f - \alpha_2 \nabla_{\theta_f} [l_c(x^s,y^s)+\lambda l_{wd}(x^s,x^t)]$
\ENDWHILE 
\end{algorithmic}
\end{algorithm}

\section{Experiments}
\label{sec:tests}
\subsection{Data description}
\label{sec:datasets}

\begin{figure}[!]
              \centering
              \includegraphics[width=1\columnwidth]{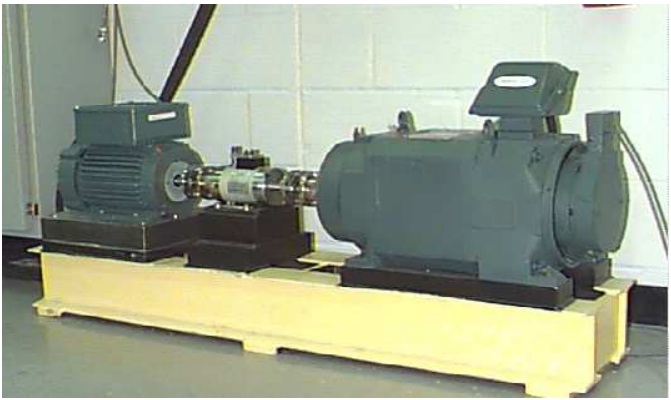}\\
              \caption{Experimental test-bed in Case Western Reserve University (CWRU) for bearing fault diagnosis.}
                \label{fig:test_bed}
  \end{figure}

To validate the effectiveness of the proposed DTL method for fault diagnosis problem, we introduce a benchmark bearing fault dataset acquired by Case Western Reserve University (CWRU) data centre. An experiment test-bed (see in Fig. \ref{fig:test_bed}) is used to conduct the signals for the detection of defects on bearings. Four types of bearing conditions are inspected, namely health condition, fault on inner race, fault on outer race, and fault on roller, and all those situations are sampled with 12\,KHz frequency. Meanwhile, each fault type are running with different level of fault severity (0.007-inch, 0.014-inch, and 0.021-inch fault diameters). Each type of faulted bearing was equipped with the test motor, which runs under four different motor speeds (i.e., 1797\,rpm, 1772\,rpm, 1750\,rpm, and 1730\,rpm). Vibration signal of each experiment was recorded for fault diagnosis.   


\textit{Data pre-processing:} Simple data pre-processing techniques are applied to the bearing datasets:

\begin{enumerate}
  \item To modify the faulty signal to a stationary process, we here divide the samples to keep each sample has 2000 measurements in both $\mathscr{D}^s$ and $\mathscr{D}^t$.
  \item Fast Fourier transform (FFT) computes the power spectrum in frequency domain of every sample.
  \item Clip the left side of the power spectrum calculated by FFT as the input for WD-DTL. Therefore, each input sample has 1000 measurements. 
 \end{enumerate}

We proposed three transfer scenarios, including two unsupervised scenarios and one supervised scenario (refer to Table \ref{tb:sum_scenarios}), they are:

\begin{table*}[]
\centering
\caption{Summarize of transfer scenarios and tasks.}
\label{tb:sum_scenarios}
\begin{tabular}{c|c|c|c|c|c|c}
\toprule
Scenario                               & \begin{tabular}[c]{@{}c@{}}Unsupervised\\ or Supervised\end{tabular} & \begin{tabular}[c]{@{}c@{}}Transfer \\ task\end{tabular} & \begin{tabular}[c]{@{}c@{}}Transfer \\ condition\end{tabular} & \begin{tabular}[c]{@{}c@{}}Training \\ \end{tabular}                                                     & \begin{tabular}[c]{@{}c@{}}Testing\\ \end{tabular}                                 & \begin{tabular}[c]{@{}c@{}}Classification\\ categorize\end{tabular}                                                                                                      \\ \hline
\multirow{12}{*}{\textbf{US-Speed}} & \multirow{12}{*}{Unsupervised}                                       & \textbf{US(A)-US(B)}                                     & 1797-1772rpm                                                  & \multirow{14}{*}{\begin{tabular}[c]{@{}c@{}}Source labeled: 100\%\\ \\ Target unlabeled: 100\%\end{tabular}} & \multirow{14}{*}{\begin{tabular}[c]{@{}c@{}}Target unlabeled:\\ 100\%\end{tabular}} & \multirow{16}{*}{\begin{tabular}[c]{@{}c@{}}Normal\\  (denoted by 0)\\ Inner Race\\ (denoted by 1)\\ Outer Race\\ (denoted by 2)\\ Roller\\ (denoted by 3)\end{tabular}} \\ \cline{3-4}
                                       &                                                                      & \textbf{US(A)-US(C)}                                     & 1797-1750rpm                                                  &                                                                                                              &                                                                                     &                                                                                                                                                                          \\ \cline{3-4}
                                       &                                                                      & \textbf{US(A)-US(D)}                                     & 1797-1730rpm                                                  &                                                                                                              &                                                                                     &                                                                                                                                                                          \\ \cline{3-4}
                                       &                                                                      & \textbf{US(B)-US(A)}                                     & 1772-1797rpm                                                  &                                                                                                              &                                                                                     &                                                                                                                                                                          \\ \cline{3-4}
                                       &                                                                      & \textbf{US(B)-US(C)}                                     & 1772-1750rpm                                                  &                                                                                                              &                                                                                     &                                                                                                                                                                          \\ \cline{3-4}
                                       &                                                                      & \textbf{US(B)-US(D)}                                     & 1772-1730rpm                                                  &                                                                                                              &                                                                                     &                                                                                                                                                                          \\ \cline{3-4}
                                       &                                                                      & \textbf{US(C)-US(A)}                                     & 1750-1797rpm                                                  &                                                                                                              &                                                                                     &                                                                                                                                                                          \\ \cline{3-4}
                                       &                                                                      & \textbf{US(C)-US(B)}                                     & 1750-1772rpm                                                  &                                                                                                              &                                                                                     &                                                                                                                                                                          \\ \cline{3-4}
                                       &                                                                      & \textbf{US(C)-US(D)}                                     & 1750-1730rpm                                                  &                                                                                                              &                                                                                     &                                                                                                                                                                          \\ \cline{3-4}
                                       &                                                                      & \textbf{US(D)-US(A)}                                     & 1730-1797rpm                                                  &                                                                                                              &                                                                                     &                                                                                                                                                                          \\ \cline{3-4}
                                       &                                                                      & \textbf{US(D)-US(B)}                                     & 1730-1772rpm                                                  &                                                                                                              &                                                                                     &                                                                                                                                                                          \\ \cline{3-4}
                                       &                                                                      & \textbf{US(D)-US(C)}                                     & 1730-1750rpm                                                  &                                                                                                              &                                                                                     &                                                                                                                                                                          \\ \cline{1-4}
\multirow{2}{*}{\textbf{US-Location}}  & \multirow{2}{*}{Unsupervised}                                        & \textbf{US(E)-US(F)}                                     & Drive End-Fan End                                             &                                                                                                              &                                                                                     &                                                                                                                                                                          \\ \cline{3-4}
                                       &                                                                      & \textbf{US(F)-US(E)}                                     & Fan End-Drive End                                             &                                                                                                              &                                                                                     &                                                                                                                                                                          \\ \cline{1-6}
\multirow{2}{*}{\textbf{S-Location}}   & \multirow{2}{*}{Supervised}                                          & \textbf{S(E)-S(F)}                                       & Drive End-Fan End                                             & \multirow{2}{*}{\begin{tabular}[c]{@{}c@{}}Source labeled: 100\%\\ Target labeled: $\sim$0.5\%\end{tabular}}       & \multirow{2}{*}{\begin{tabular}[c]{@{}c@{}}Target unlabeled:\\ 25\%\end{tabular}}   &                                                                                                                                                                          \\ \cline{3-4}
                                       &                                                                      & \textbf{S(F)-S(E)}                                       & Fan End-Drive End                                             &                                                                                                              &                                                                                     &                                                                                                                                                                          \\ \bottomrule
\end{tabular}
\end{table*}

\begin{enumerate}
  \item Unsupervised transfer between motor speeds \textbf{(US-Speed)}: For this scenario, we test the data with 12\,KHz sampling frequency acquired at the drive end of the motor, and ignore the level of fault severities. Thus, we construct 4-way classification tasks (i.e., health condition, and three fault conditions with faults on inner race, outer race and roller), across 4 domains with different motor speeds: 1797\,rpm $\textbf{(US(A))}$, 1772\,rpm $\textbf{(US(B))}$, 1750\,rpm $\textbf{(US(C))}$, and 1730\,rpm $\textbf{(US(D))}$. In total, for this scenario, we evaluate our proposed method over 12 transfer tasks.  
  \item Unsupervised transfer between datasets at two sensor locations \textbf{(US-Location)}: For this scenario, we focus on domain adaptation between  different sensor locations but ignore the level of fault severities and the differences in motor speeds. Again, we construct 4-way classification tasks for health and three fault conditions, across 2 domains (2 tasks) where vibration acceleration data acquired by two sensors placed at the drive end $\textbf{(US(E))}$ and fan end $\textbf{(US(F))}$ of the motor housing respectively.
 \item Supervised transfer between datasets at two sensor locations \textbf{(S-Location)}: this scenario uses the same settings as the previous scenario \textbf{US-Location}, except for the specified change of adding a small amount of labeled data ($\sim 0.5\%$) of target domain in source domain which aims to enhance the classification performance.
\end{enumerate}

To evaluate the efficiency of our proposed approach WD-DTL on bearing fault diagnosis problem, other approaches are also tested on the same dataset for comparison purpose:

\begin{itemize}
  \item \textit{CNN (no transfer):} This model is the pre-trained network described in Section \ref{sec:FE}, which is trained based on the labeled source data and applied to test the classification result on the target domain directly. 
  \item \textit{DAN:} We follow the idea in work \cite{long2015learning}, of which proposed a deep adaptation  network (DAN) for learning transferable features via MK-MMD in deep neural networks. 
The MMD metric is an integral probability metrics which measures the distance between two probability distributions via mapping the samples into a Reproducing Kernel Hilbert Space (RKHS).  Domain adaptation via MMD has been explored for image classification in several works, see in \cite{sejdinovic2013equivalence, gretton2012kernel, long2015learning}.
   \item In addition, to evaluate the feature extraction ability of CNN compared to the use of conventional statistical features. Results of traditional transfer learning methods using statistical (handcrafted) features \cite{guo2018deep}, including transfer component analysis \textit{(TCA)} \cite{pan2011domain}, joint distribution adaptation \textit{(JDA)} \cite{long2013transfer}, and CORrelation ALignment \textit{(CORAL)} \cite{sun2016return}, are also provided for comparison. 
\end{itemize}

This work will mainly focus on the comparison between those deep transfer learning methods (DAN and WD-DTL) and CNN.

\subsection{Implementation details}
\label{sec:Imp_details}

TensorFlow \cite{abadi2016tensorflow} is used as software framework for all our experiments using deep learning flow, and those models are all trained with Adam optimizer. We test each approach for five times over 5000 iterations and record the best result of each test. We take the averages and 95\% confidential interval of classification accuracy for comparison. The sample size for motor speed tasks $\textbf{(A)}$, $\textbf{(B)}$, $\textbf{(C)}$, and $\textbf{(D)}$ are 1026, 1145, 1390, and 1149 respectively. The sample size for different sensor location tasks $\textbf{(E)}$ and $\textbf{(F)}$ are 3790 and 4710 respectively. The batch size $n$ is fixed as 32 for all experiments. 

 $\textbf{CNN}$: Our CNN architecture is comprised of two convolutional layers ($\textit{Conv1-Conv2}$), two max-pooling layers ($\textit{Pool1-Pool2}$), and two fully-connected layers ($\textit{FC1-FC2}$). The activation function in output layer is $\textit{Softmax}$ while $\textit{ReLu}$ is used in convolutional layers. The neuron number in $\textit{FC1}$ and $\textit{FC2}$ are 128 and 4, respectively. Filters, kernel size, and stride of each layer can refer to Table \ref{tb:CNNpara}. Before transfer, we fine-tune the CNN models which achieve their best validation  accuracies for all transfer scenarios.

\begin{table}[ht!]
\caption{Parameters in the CNN model}
\label{tb:CNNpara}
\center
\begin{tabular}{lcccc}
\hline
Layer       & Filters & Kernel size & Stride\\ \hline
\hline
Conv1       & 8     & 1x20  & 2            \\
Pool1 & -    & 1x2   & 2                \\
Conv2       & 16     & 1x20   & 2              \\
Pool2 & -    & 1x2    & 2              \\
\hline
\end{tabular}
\end{table}

$\textbf{DAN}$:  The convolutional layers ($\textit{Conv1-Conv2}$) of the CNN network is used to be the feature extractor. Then, to minimize the domain distance between the source and target domains, FC1 is used as the hidden layer for adaptation. The final representations of the hidden layer in both domains are embedded to RKHS to reduce the MK-MMD distance. The final objective function is the combination of the MK-MMD loss and  the classification loss. Best classification accuracies are obtained for transfer scenarios by tuning the balancing coefficient for the discrepancy loss. 
 
  $\textbf{WD-DTL}$: WD-DTL method has been summarized in Fig. \ref{fig:WD-DTL} and Algorithm \ref{code:alg1}. Similar to DAN, convolutional layers (Conv1-Conv2) are used to extract features. The nodes of hidden layers in the domain critic network are set to 128 and 1, respectively. The training step $C$ is set to 10. The learning rates for the discriminator and the domain critic are $\alpha_1=10^{-3}$ and $\alpha_2=2\times 10^{-4}$ respectively. The gradient penalty $\rho$ is set to 10. Balance coefficient $\lambda$ for optimizing the minimum operator is 0.1 and 0.8 for motor speed transfer and sensor location transfer, respectively. 
  
 In terms of the traditional transfer learning methods TCA, JDA and CORAL, the regularization term $\lambda$ is chosen from \{0.001 0.01 0.1 1.0 10 100\}. SVM is used in TCA and CORAL for classification. 

\subsection{Results and discussion}
\label{sec:comparison}
The results of transfer tasks for WD-DTL and the other two approaches are shown in Table \ref{tb:result}. For the transfer task with unlabeled data set in target domain (i.e., scenario \textbf{US-Speed} and \textbf{US-Location}), we can observe that WD-DTL significantly outperforms CNN with a large margin, which achieves approximately 13.6\% and 25\% increases in average accuracies for motor speed and sensor location transfer tasks, respectively. In addition, the WD-DTL transfer accuracies are better than most of the DAN results (average 5\% increase), except transfer task $\textbf{US(D)}\rightarrow \textbf{US(A)}$ which result in less than 1\% accuracy difference. 

To summarize the results, we can make the following observations: 1) WD-DTL achieves the best transfer accuracies with 95.75\% average score, confirming the effectiveness of Wasserstein distance in learning transferable features using CNN-based model; 2) Without domain adaptation, CNN method already has the ability to achieve good classification performance for the motor speed transfer tasks, due to its excellent feature detection ability; 3) The accuracies of CNN, DAN and WD-DTL on transfer tasks of scenario \textbf{US-Location} are not better than the transfer tasks of scenario \textbf{US-Speed}, due to the characteristics of signals obtained at different sensor location (Fan End and Drive End) are more different than the difference between motor speeds; and 4) The proposed WD-DTL approach shows a good ability to solve supervised problem with a small number of labeled data. Supervised transfer tasks \textbf{S(E)$\rightarrow$ S(F)} and \textbf{S(E) $\rightarrow$ S(F)} are carried out using only 0.5\% sample size of the unsupervised case, but achieve as good as performance compared to the unsupervised case which using 100\% unlabeled sample. Further analysis of the effect of sample size for both supervised and unsupervised transfer learning will be shown in Section \ref{sec:dataset_acc}.

\begin{table*}[h!]
\caption{Performance of transfer tasks (Accuracy \%)}
\label{tb:result}
\centering
\begin{tabular}{@{}ccccccc@{}}
\toprule
          & \textbf{TCA} & \textbf{JDA} & \textbf{CORAL} & \textbf{CNN} & \textbf{DAN} & \textbf{WD-DTL} \\ \midrule
\textbf{US(A)$\rightarrow$US(B)} & 26.55 & 65.07 ($\pm$ 7.55) & 59.18 & 82.75 ($\pm$ 6.77)      & 92.97   ($\pm$ 3.88)     & \textbf{97.52}  ($\pm$ 3.09)         \\
\textbf{US(A)$\rightarrow$US(C)} & 46.80 & 51.31 ($\pm$ 1.56) & 62.14  & 78.65  ($\pm$ 4.54)       & 85.32   ($\pm$ 5.26)     & \textbf{94.43}    ($\pm$ 2.99)       \\
\textbf{US(A)$\rightarrow$US(D)} & 26.57 & 57.70 ($\pm$ 8.59) & 49.83   & 82.99  ($\pm$ 5.89)       & 89.39    ($\pm$ 4.37)   & \textbf{95.05}  ($\pm$ 2.12)         \\
\textbf{US(B)$\rightarrow$US(A)} & 26.63 & 71.19 ($\pm$ 1.21) & 53.57 & 84.14  ($\pm$ 6.63)       & 94.43    ($\pm$ 2.95)    & \textbf{96.80}  ($\pm$ 1.10)        \\
\textbf{US(B)$\rightarrow$US(C)} & 26.60 & 69.80 ($\pm$ 5.67) & 57.28 & 85.41   ($\pm$ 9.44)      & 90.43  ($\pm$ 4.62)      & \textbf{99.69}    ($\pm$ 0.59)       \\
\textbf{US(B)$\rightarrow$US(D)}  & 26.57 & 88.50 ($\pm$ 1.96) & 60.53 & 86.09  ($\pm$ 4.63)       & 87.37  ($\pm$ 5.42)        & \textbf{95.51}  ($\pm$ 2.52)        \\
\textbf{US(C)$\rightarrow$US(A)}  & 26.63 & 56.42 ($\pm$ 2.52) & 54.03 & 76.50   ($\pm$ 3.76)      & 89.88  ($\pm$ 1.57)     & \textbf{92.16}  ($\pm$ 2.61)          \\
\textbf{US(C)$\rightarrow$US(B)} & 26.66 & 69.18 ($\pm$ 1.90) & 76.66 & 82.75   ($\pm$ 5.51)      & 92.93    ($\pm$ 1.57)    & \textbf{96.03}  ($\pm$ 6.27)          \\
\textbf{US(C)$\rightarrow$US(D)} & 46.75 & 77.45 ($\pm$ 0.83) & 70.34 & 87.04   ($\pm$ 6.81)      & 90.66    ($\pm$ 5.24)    & \textbf{97.56}   ($\pm$ 3.31)        \\
\textbf{US(D)$\rightarrow$US(A)} & 46.74 & 61.72 ($\pm$ 5.48) & 59.78 & 79.23     ($\pm$ 6.96)    & \textbf{90.88}  ($\pm$ 1.82)      & 89.82   ($\pm$ 2.41)       \\
\textbf{US(D)$\rightarrow$US(B)} & 46.79 & 74.03 ($\pm$ 0.86) & 59.73 & 79.73    ($\pm$ 5.49)     & 87.91  ($\pm$ 2.42)      & \textbf{95.16}  ($\pm$ 3.67)         \\
\textbf{US(D)$\rightarrow$US(C)} & 26.60 & 65.24 ($\pm$ 4.18) & 63.02 & 80.64  ($\pm$ 4.23)      & 92.94   ($\pm$ 3.96)     & \textbf{99.62}  ($\pm$ 0.80)        \\ 
\textbf{Average} & 33.32 & 67.35 ($\pm$ 3.53) & 56.01 & 82.10 ($\pm$ 5.89)     & 90.42 ($\pm$ 3.59)    & \textbf{95.75} ($\pm$ 2.62)      \\
\midrule
\textbf{US(E)$\rightarrow$US(F)} & 19.05 & 57.35 ($\pm$ 0.47) & 47.97 &   39.07   ($\pm$ 2.22)      &    56.89   ($\pm$ 2.73)        &   \textbf{64.17}   ($\pm$ 7.16)            \\
\textbf{US(F)$\rightarrow$US(E)} & 20.45 & \textbf{66.34} ($\pm$ 4.47) & 39.87 &      39.95  ($\pm$ 3.84)       &    55.97    ($\pm$ 3.17)       &    64.24   ($\pm$ 3.87)              \\ 
\textbf{Average} & 19.75 & 61.85 ($\pm$ 2.47) & 43.92& 39.51 ($\pm$ 3.03)     & 56.43 ($\pm$ 2.95)    & \textbf{64.20} ($\pm$ 5.52)      \\\midrule
\textbf{S(E)$\rightarrow$S(F)} & 20.43 & 65.48 ($\pm$ 0.57) & 51.77 &    54.04   ($\pm$ 7.67)      &    59.68   ($\pm$ 4.61)        &   \textbf{65.69}   ($\pm$ 3.74)            \\
\textbf{S(F)$\rightarrow$S(E)} & 19.02 & 59.07 ($\pm$ 0.56) & 47.88 &      50.47  ($\pm$ 5.74)       &    58.78   ($\pm$ 5.67)       &    \textbf{64.15}   ($\pm$ 5.52)              \\ 
\textbf{Average} & 19.73 & 62.28 ($\pm$ 0.57) & 49.83& 52.26 ($\pm$ 6.71)     & 59.23 ($\pm$ 5.14)    & \textbf{64.92} ($\pm$ 4.63)      \\\bottomrule
\end{tabular}
\end{table*}


\subsection{Empirical Analysis}
\label{sec:Empirical}

\subsubsection{\textbf{Feature visualization}}
\label{sec:Visual}

\begin{figure*}[htb!]
\centering
  \includegraphics[width=2\columnwidth]{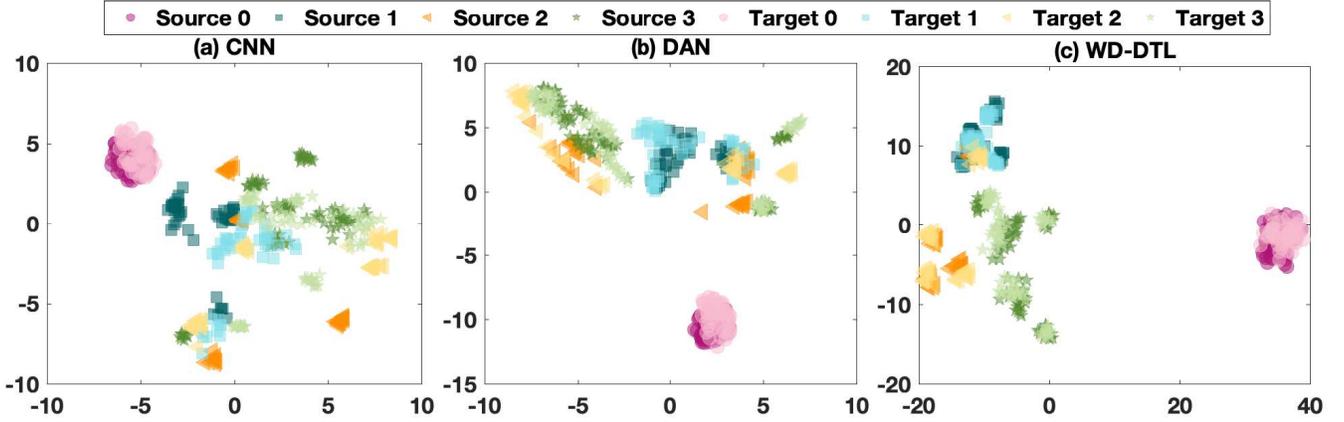}
  \caption{Network visualization revealed by t-SNE embeddings of transfer task  $\textbf{US(C)}\rightarrow\textbf{US(A)}$ with: (a) CNN approach, (b) DAN approach, and (c) WD-DTL approach. t-SNE is applied on the features in the last layer assigned by CNN-based feature extractor network, for both source and target domains. Four colors/shapes represent four conditions, namely normal condition, fault on inner race, fault on outer race, and fault on roller (with corresponding labels 0-3).}
  \label{fig:t-sne}
\end{figure*}

\begin{figure*}[htb!]
\centering
  \includegraphics[width=2\columnwidth]{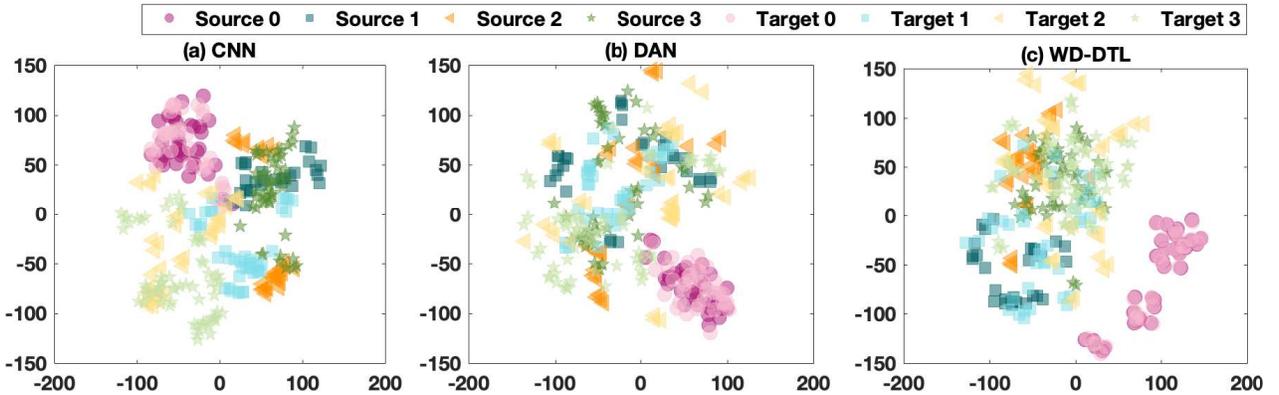}
  \caption{Network visualization of transfer task  $\textbf{US(E)}\rightarrow\textbf{US(F)}$ with: (a) CNN approach, (b) DAN approach, and (c) WD-DTL approach.}
  \label{fig:t-sne_location}
\end{figure*}

To further evaluate the transfer performance of the proposed WD-DTL framework, t-distributed stochastic neighbor embedding (t-SNE) is employed to perform the nonlinear dimensionality reduction for network visualization. For comparison purpose, CNN and DAN transfer results for same tasks are also presented. 

For transfer tasks between motor speeds, i.e., scenario \textbf{US-Speed}, we randomly choose task $\textbf{US(C)}\rightarrow\textbf{US(A)}$ to visualize the learned feature representations under different motor speeds. Fig. \ref{fig:t-sne} shows the comparison results. It can be observed that the clusters in Fig. \ref{fig:t-sne}(c) formed by our proposed WD-DTL are better separated than the CNN network result in Fig. \ref{fig:t-sne}(a) that was not trained for domain adaptation and the DAN domain adaptation result in Fig. \ref{fig:t-sne}(b). For example, in Fig. \ref{fig:t-sne}(a) with CNN approach, three types of fault features are inspected with large overlapped areas, and some outer-race faults (yellow color with label 2)  fall into other fault types. Similarly, in Fig. \ref{fig:t-sne}(b) with DAN approach, outer-race faults is also hardly be separated from other fault types. With our WD-DTL approach, four conditions are clearly separated into different clusters. More importantly, we can observe the obvious improvement of domain adaptation due to source and target domain features are almost mixed into the same cluster.

For transfer tasks between different sensor locations, i.e., scenarios \textbf{US-Location} and \textbf{S-Location}, t-SNE results of transfer task $\textbf{US(E)}\rightarrow\textbf{US(F)}$ are in Fig. \ref{fig:t-sne_location} provided. It can be viewed that even WD-DTL shows better clustering result than CNN and DAN, faults types 1, 2, and 3 are hard to be separated clearly into individual clusters. It must be emphasized that above results are carried out by using 100\% (4710) sample size in target domain, and even in this case the performance is not satisfied enough. This raise the problem of how to enhance the transfer learning performance when signals in source and target domains are relevant but not similar enough. We investigate this problem in the next subsection.

\subsubsection{\textbf{Effect of sample size on unsupervised and supervised accuracy}}
\label{sec:dataset_acc}

\begin{figure*}[!]
\centering  \includegraphics[width=1.7\columnwidth]{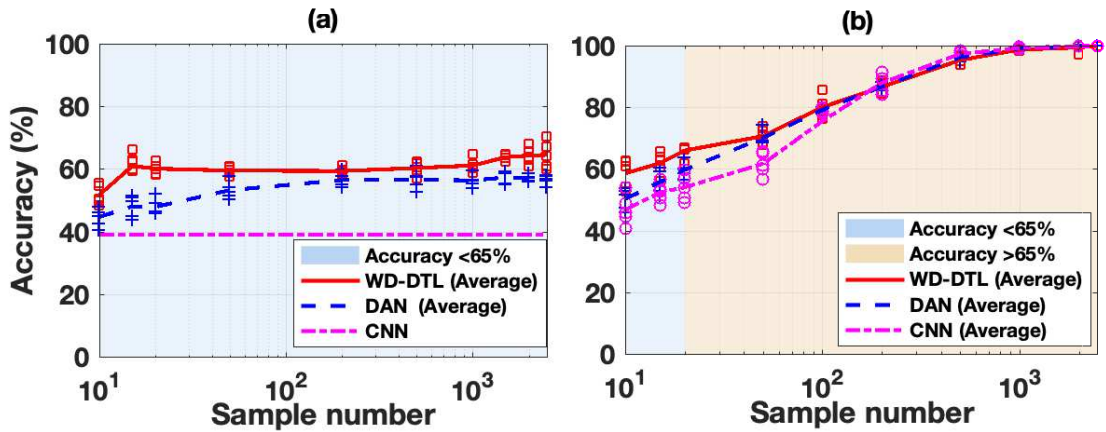}
  \caption{Accuracy variation curve of task (a) $\textbf{US(E)}\rightarrow \textbf{US(F)}$ and (b) $\textbf{S(E)}\rightarrow \textbf{S(F)}$, where sample number is increased from 10 to 2500. Red, blue, and purple lines present the evolution of the average accuracy of five times' results under different sample number.}
  \label{fig:sample_size}
\end{figure*}

Next, we investigate the influence of data size on transfer task accuracy for our proposed method WD-DTL. For each sample number tested, same experiment is repeated five times and transfer learning accuracies are recorded. As it has known that our propose WD-DTL method already achieved very good performance (average 95.75\% accuracy in Table \ref{tb:result}) for unsupervised transfer scenario \textbf{US-Speed}. Fig. \ref{fig:sample_size} displays the accuracy variation curve for WD-DTL of tasks $\textbf{US(E)}\rightarrow \textbf{US(F)}$ and $\textbf{S(E)}\rightarrow \textbf{S(F)}$ with respect to scenario \textbf{US-Location} and \textbf{S-Location}. Diagnosis accuracies will be saturated around a fixed value when sample number larger than 2500, thus we only show the result from 10 to 2500.

\begin{figure*}[htb!]
\centering \includegraphics[width=2\columnwidth]{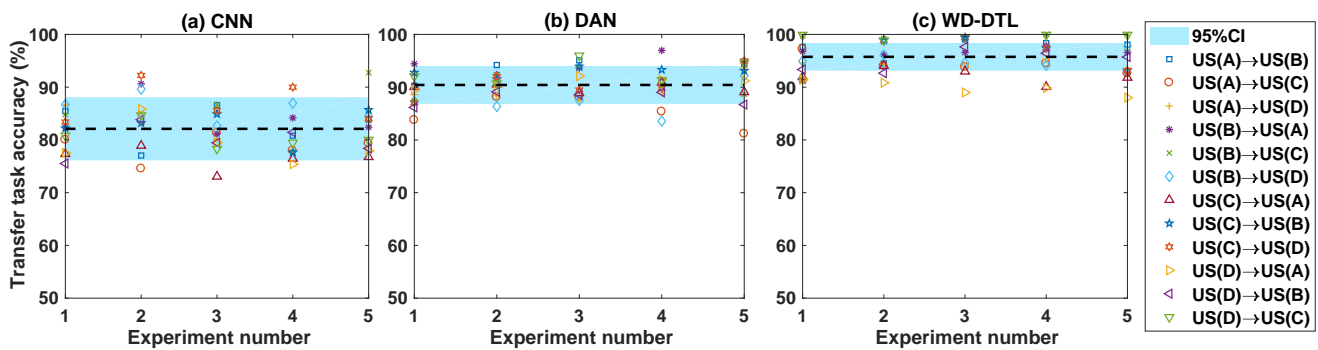}
  \caption{Experiment number vs. transfer task accuracy of motor speed transfer tasks, for (a) CNN approach, (b) DAN approach, and (c) WD-DTL approach. Black dotted lines represent the average scores. Blue zone is the 95\% confidential interval of each approach.}
  \label{fig:robust}
\end{figure*}

In Fig. \ref{fig:sample_size}(a), it can be observed that the accuracy of WD-DTL is increased from 59.47\% and the final test accuracy is confined around 64\%. While the sample number is increasing, fault diagnosis accuracies of WD-DTL approach are all higher than DAN and CNN. This analysis reveals that, for this unsupervised scenario, the increase of sample number could improve the transfer learning accuracy, however, the improvement is limited (less than 5\%) even with 100\% sample number in target domain. To solve this problem, in Fig. \ref{fig:sample_size}(b), we employ a small amount of labeled data to improve the fault diagnosis accuracy, which is associated with the case with limited labeled data in real industry application. The plot shows that when the labeled sample size larger than 20 of 4710 the transfer learning accuracy of WD-DTL will surpass the case in Fig. \ref{fig:sample_size}(a) with 100\% sample size (blue zone in Fig. \ref{fig:sample_size}(a)). More specifically, only using 100 labeled sample, (equivalent to 25 for each fault categorization) could achieve 80\% transfer learning accuracy, indicating our proposed WD-DTL is also an optimal framework for supervised transfer task.

Based on the above discussions, we hereby offer two solutions for manufacturers of using the proposed WD-DTL approach: 1) when facing the transfer tasks between similar signals in source and target domains, such as transfer learning between different motor speeds, unsupervised transfer learning with unlabeled data is enough to obtain very good fault diagnosis accuracy (larger then 95\%); and 2)  when facing the transfer tasks between relevant signals but not similar enough, such as transfer learning between different  sensor locations, a small amount of labeled sample will greatly improve the transfer learning accuracy compared to the unsupervised case with large amount of unlabeled sample data.

\subsubsection{\textbf{Algorithm robustness evaluation}}
\label{sec:stab}

The robustness of our proposed algorithm WD-DTL is investigated and compared with CNN and DAN approaches. We run each task for five times and store the transfer accuracy of each task. Fig. \ref{fig:robust} gives an illustration of the variation of transfer task accuracy on 12 tasks of motor speed transfer scenario. We can observe that not only the WD-DTL accuracy is higher than other two approaches but also it has a narrower 95\% confidential interval than other two approaches. This confirms our motivation of using CNN-based network and Wasserstein distance for domain adaptation, since both the accuracy and model robustness of feature transferability is enhanced by using our proposed algorithm. 

During our experiments, we also found that the robustness of transfer model for mechanical system is worse than image classification transfer model, which might due to the large noise in the acquired acceleration signals. In our future work, this might can be solved by employing some basic signal processing techniques to filter the noise.

\section{Conclusion}
\label{sec:conclusion}
To achieve intelligent fault diagnosis, we proposed a novel Deep Transfer Learning architecture via Wasserstein Distance (WD-DTL) to enhance the domain adaptation ability. WD-DTL is constructed based on a deep learning flow (CNN architecture) to extract features and introduces a domain critic to learn domain invariant feature representations. Through an adversarial training process, WD-DTL significantly reduce the domain discrepancy thanks to its gradient property of Wasserstein distance over other state-of-the-arts distances and divergences. Our proposed method is tested on a CRWU benchmark bearing fault diagnosis dataset and compared with the base CNN model, DAN metric and other traditional transfer learning methods over 16 transfer tasks. Performance of all the transfer tasks demonstrate that WD-DTL outperforms other approaches with much better classification accuracies. Empirical results also show that 1) our proposed method achieves higher robustness for motor speed transfer tasks, and 2) WD-DTL is a novel approach which could contribute to solve both unlabeled and insufficient labeled data problems in real industry applications. Future work includes investigating more transfer scenarios (e.g. transfer learning between different machines) for intelligent fault diagnosis and optimizing the architecture of our proposed algorithm.



\end{document}